\documentclass[10pt,twocolumn,letterpaper]{article}

\usepackage{cvpr}              % To produce the CAMERA-READY version
\usepackage{graphicx}
\usepackage{amsmath}
\usepackage{amssymb}
\usepackage{booktabs}

\usepackage{color}
\usepackage{colortbl}
\usepackage{multirow}

\makeatletter
\newcommand\footnoteref[1]{\protected@xdef\@thefnmark{\ref{#1}}\@footnotemark}
\makeatother

\newcommand{\shorteq}{%
  \settowidth{\@tempdima}{-}% Width of hyphen
  \resizebox{\@tempdima}{\height}{=}%
}

\usepackage{amssymb,fge}

\usepackage{amsmath, amssymb}
\usepackage{subcaption}
\usepackage[utf8]{inputenc}
\usepackage[T1]{fontenc}
\usepackage[linesnumbered,ruled,vlined]{algorithm2e}
\usepackage{graphicx}
\usepackage{amsfonts}
\usepackage{enumitem}
\usepackage{soul}
\usepackage{hhline}
\usepackage{multirow, makecell}
\usepackage{float}
\usepackage{booktabs}

\usepackage{amsthm}
\usepackage{color}
\usepackage{transparent}
\usepackage{footmisc}
\usepackage{setspace}
\usepackage{textcomp}
\usepackage{mathtools}

\theoremstyle{plain}

\mathchardef\mhyphen="2D

\definecolor{RowColorCode}{rgb}{0.9,0.57,0.95}

\usepackage[pagebackref,breaklinks,colorlinks]{hyperref}

\usepackage[capitalize]{cleveref}
\crefname{section}{Sec.}{Secs.}
\Crefname{section}{Section}{Sections}
\Crefname{table}{Table}{Tables}
\crefname{table}{Tab.}{Tabs.}

\def\cvprPaperID{28} % *** Enter the CVPR Paper ID here
\def\confName{CVPR}
\def\confYear{2024}

\begin{document}

%\title{Beyond Single-Concept Customization of Video Generation Models}
\title{Text Prompting for Multi-Concept Video Customization by Autoregressive Generation} %kihyuks
% \title{Multi-Concept Personalization of Video Generation Model} %kihyuks

%Oct25:MBZ: Not a big fan of "Causality is All You Need?" part. I know it's catchy but it's also too much on the nose :) Done !

\author{Divya Kothandaraman$^{1,2}$, Kihyuk Sohn$^{3}$, Ruben Villegas$^{2}$, Paul Voigtlaender$^{3}$, \\ Dinesh Manocha$^{1}$, Mohammad Babaeizadeh$^{2}$ \\
University of Maryland College Park$^{1}$, Google DeepMind$^{2}$, Google Research$^{3}$\\
%\and
}

%\maketitle
\twocolumn[{%
\renewcommand\twocolumn[1][]{#1}%
\maketitle
\begin{center}
    \centering
    \captionsetup{type=figure}
    \includegraphics[width=0.9\textwidth]{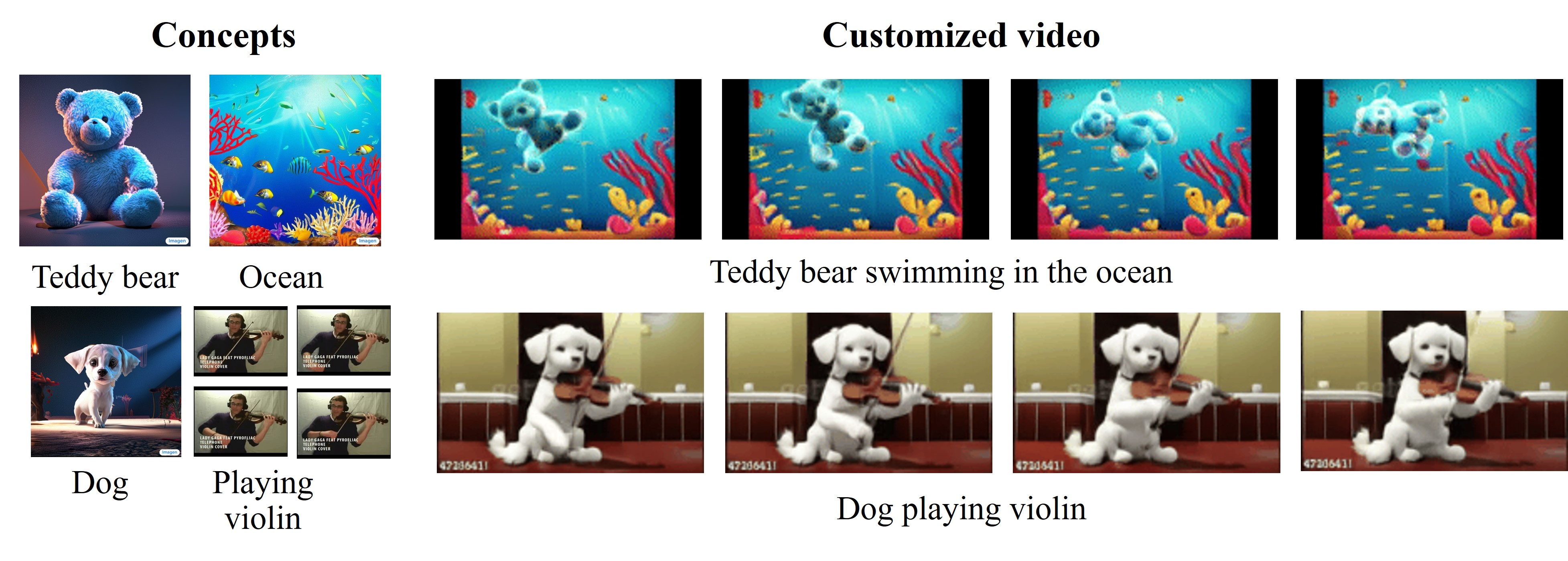}
    % \captionof{figure}{We present a method for multi-concept customization of text-to-video models. Our method uses just the pre-trained model and concept-specific image(s)/ video(s) to customize concepts, and does not require access to any additional data, to generate videos such as a teddy bear swimming in the ocean and a dog playing violin.}
    \vspace{-6pt}
    \captionof{figure}{We present a method for multi-concept customization of text-to-video models. Our method only relies on a pre-trained text to video model and concept specific data in the form of image(s) or video(s). In the illustration above, our model uses an image of a teddy bear and ocean background (top row) and an image of a dog and video of playing violin (bottom row) to generate the customized videos.}
\end{center}%
}]

\begin{abstract}

%Personalization methods like DreamBooth are successful in generating videos of custom subjects, however, more often than not, they are unable to generate videos containing more than one custom concepts. In this paper, 
We present a method for multi-concept customization of pretrained text-to-video (T2V) models. Intuitively, the multi-concept customized video can be derived from the (non-linear) intersection of the video manifolds of the individual concepts, which is not straightforward to find. We hypothesize that sequential and controlled walking towards the intersection of the video manifolds, directed by text prompting, leads to the solution. To do so, we generate the various concepts and their corresponding interactions, sequentially, in an autoregressive manner. Our method can generate videos of multiple custom concepts (subjects, action and background) such as a teddy bear running towards a brown teapot, a dog playing violin and a teddy bear swimming in the ocean. We quantitatively evaluate our method using videoCLIP and DINO scores, in addition to human evaluation. Videos for results presented in this paper can be found at https://github.com/divyakraman/MultiConceptVideo2024. 
\footnote{Work done while Divya was an intern at Google DeepMind.}

\end{abstract}

%{\color{red}please change LoRA to adapter at all places. - done}

\section{Introduction}

Text-to-video generation models such as Video Diffusion Models~\cite{ho2022video}, Make-A-Video~\cite{singer2022make}, Imagen Video~\cite{ho2022imagen}, Phenaki~\cite{villegas2022phenaki} CogVideo~\cite{hong2022cogvideo}, VideoFusion~\cite{luo2023videofusion}, Gen-1~\cite{esser2023structure} are capable of generating visually appealing videos. However, their application is constrained due to the inability to generate long videos (beyond a few seconds)~\cite{yin2023nuwa,harvey2022flexible} and limited control~\cite{chen2023videodreamer}. These limitations are problematic mainly because when generating a long video that includes multiple subjects, we expect the same subjects to appear later in the video. For instance, if we generate a video of a specific \textit{Teapot} boiling under a specific \textit{Tree}, later in the video we expect to see the same \textit{Teapot} and the same \textit{Tree} and not another randomly generated one. One way to achieve such results with the current text to video models is via providing extreme level of details in the captions, so that the model generates the same \textit{Tree} and the same \textit{Teapot} everytime. However, this is usually impractical~\cite{ruiz2023dreambooth,ho2022video} because the models may not be sensitive to all the details and the concepts may be personal and unseen by the model during large-scale pre-training~\cite{singer2022make,ho2022video,villegas2022phenaki}. 

Another way of getting consistent subjects in consecutive video frames is that video generation should have the capability to generate clips conditioned on one or more given subjects that may or may not be a part of the training data. Model customization~\cite{ruiz2023dreambooth,gal2022image} methods such as DreamBooth~\cite{ruiz2023dreambooth} and DreaMix~\cite{molad2023dreamix} work well for single concept video customization such as generating a random video of a customized subject or making a random subject perform a custom action. However, they are not as effective in the multi-concept customization regime~\cite{chen2023videodreamer}. We hypothesise that this is due to the inability of the model in understanding the interactions between different custom concepts, that it may have not encountered during training~\cite{kumari2023multi}. Also, learning characteristics of concepts from limited data per concept results in severe overfitting~\cite{han2023svdiff}. Hence, the variance of the model w.r.t. the custom concept is low and it is predominant to use the variance of the model to overcome bias issues w.r.t. the custom concept. Additionally, when multiple concepts are customized in a video, there is a mixup of their corresponding attributes~\cite{liu2022compositional} 
%Moreover, more often than not, we have access to only a single image~\cite{kawar2023imagic,zhang2023sine} for a custom concept and wish to generate personalized videos integrating the concept seamlessly into the video scene. Additionally, the personalization may not be limited to a single custom concept, rather, we may wish to generate videos with multiple customized concepts~\cite{han2023svdiff,kumari2023multi}.  

%\begin{figure}
%    \centering
%    \includegraphics[scale=0.27]{Figures/dreamboothlora_singleconcept1.png}
%    \caption{Customization methods such as DreamBooth LoRA~\cite{ruiz2023dreambooth}, adapted to video customization, are able to successfully perform single-concept personalization such as subject personalization (`a blue teddy bear walks into the kitchen and sits on the dining table') and motion customization (`a teddy bear playing tennis in a tennis court').}
%    \label{fig:dreamboothlora_singleconcept}
%\end{figure}

%\begin{figure}
%    \centering
%    \includegraphics[scale=0.27]{Figures/dreambooth_multiconceptdrawbacks.png}
%    \caption{Customization methods such as DreamBooth LoRA~\cite{ruiz2023dreambooth} are not able to perform multi-concept customization due to the inability to generate interactions between different custom concepts caused by generalization issues w.r.t. the dataset the pretrained model was trained on, bias issues w.r.t. the image(s) or video(s) corresponding to the custom concepts, and mix-up of attributes. }
%    \label{fig:dreambooth_multiconceptdrawbacks}
%\end{figure}

Prior work on multi-concept customization of text to image models~\cite{han2023svdiff,kumari2023multi} and concurrent work on multi-concept video customization~\cite{chen2023videodreamer} are able to put together multiple subjects in the scene in a realistic manner. However, they often train individual adapter models~\cite{kumari2023multi} for each concept before projecting to a common space, making it memory inefficient for video models, and also rely on expensive data augmentation from the original training dataset. Moreover, the text-to-image customization methods suffer from attribute binding problems and artifacts~\cite{chen2023videodreamer} when extended to the video space. 

\paragraph{Main contributions.} In this paper, we present a simple, yet effective solution for multi-concept video customization based on ``teaching'' the model interactions between various concepts one-by-one in order to generate the custom video. %\mbz{can be removed or moved to the Method section: For example, consider the same two subjects as before: \textit{Teapot} and \textit{Tree}. The manifold of ``Teapots" contains variations of teapots with our specific \textit{Teapot} in it. Similarly, the manifold of "trees`` contains variations of videos with \textit{Tree}. The goal video of \textit{Teapot boiling under Tree} lies at the intersections of these manifolds, which is difficult to find and reach.}
We show that sequential and controlled traversal through the manifolds of each subject (each understood by the model) towards the intersection space of the manifolds, while generating video frames sequentially, leads to the generation of the custom video. The main intuition behind our proposed method is that while sequentially generating the frames of a video, the model needs to remember what it previously generated and synthesize the interactions accordingly while staying causal in time. Our solution for multi-concept video customization exploits the inherent strong causality in autoregressive T2V models to generate the interactions between the different concepts in the video. %\mbz{can be removed: Adding complexity sequentially rather than at once makes it easier for the model to generate the video}. 
Moreover, at each stage of generation after the first step, the availability of prior knowledge through previously generated frames helps in increasing the variance of the model, which is useful for generating the interactions between various concepts. 

We qualitatively evaluate our method on diverse scenarios of generating multi-concept customized videos such as \textit{a dog playing violin, a dog and a teddy bear eating together and a teddy bear walking in a living room}; and report quantitative metrics such as videoCLIP and DINO, along with human evaluation. We show improved results over baselines such as DreamBooth style finetuning. We also show the usefulness of our method for single concept customization in cases where compositionality is desired.

%\mbz{Nov14: this paragraph is really out of place. Consider moving it or completely removing it.} Any video contains different kinds of concepts interacting with each other. This could be a custom subject performing a custom action (such as a dog playing a violin) or two subjects interacting with each other (such as a custom dog and a custom teddy bear eating together) or subject performing a action in a background (such as a custom teddy bear walking in a custom living room). We analyze our method on these various combinations in detail. Moreover, our method is not useful just for multi concept customization, but also for single concept customization where compositionality is desired. 

\section{Related work}

\begin{figure*}
    \centering
    \includegraphics[scale=0.39]{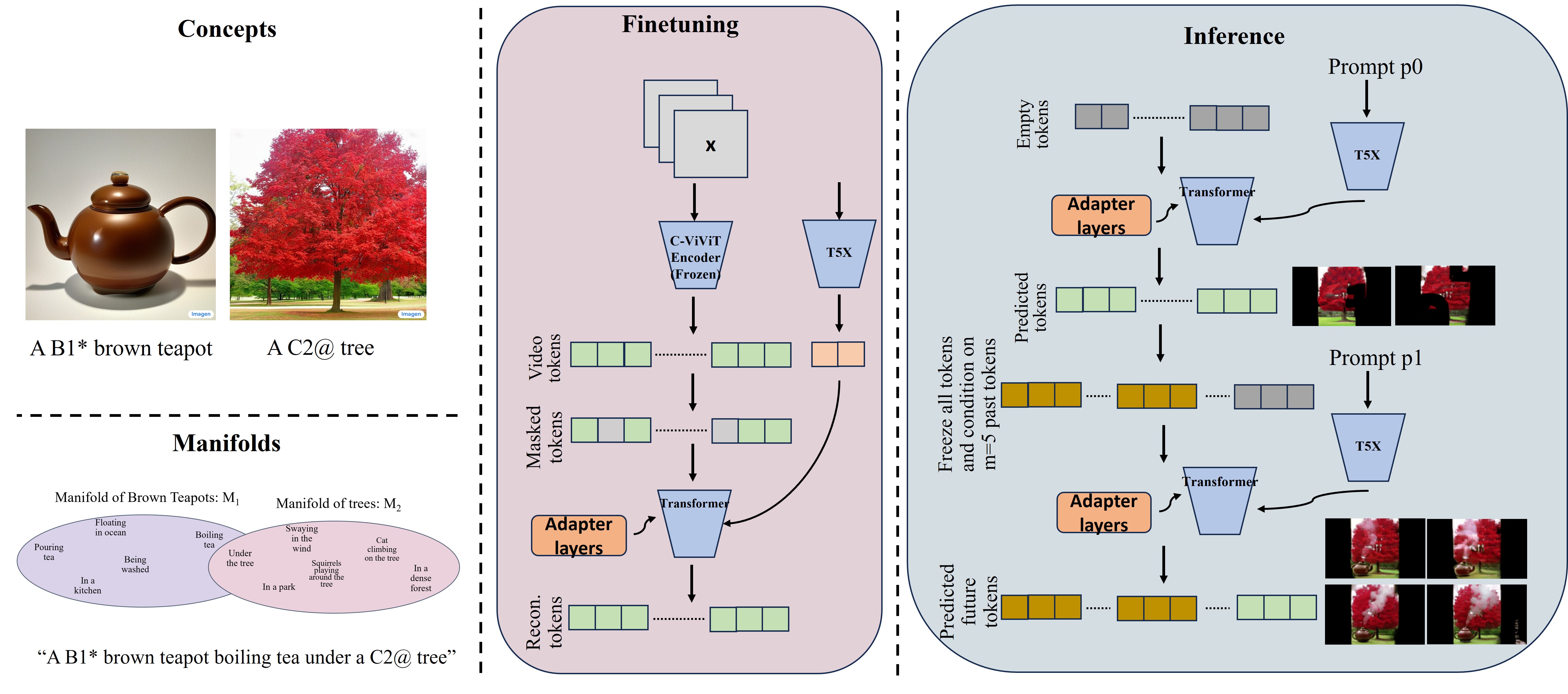}
    \caption{Proposed method for multi-concept video customization. First, we add adapter layers to the transformer architecture in an autoregressive T2V model and finetune these additional layers on the given images or videos associated with the $N$ custom concepts and their corresponding text prompts. The goal is to find the solution at the intersection of the video manifolds corresponding to various custom concepts. Then, we condition on $m$ ($=5$) prior frames and sequentially generate the custom concepts and their interactions in a controlled manner using a set of prompts $p_{0} ... p_{N}$. The prompts $p_{0} ... p_{N}$ are designed to represent the scene in a top-down manner, each prompt adding a custom concept and the associated interaction.}
    \label{fig:overview}
\end{figure*}

\subsection{Text to Video Models}

Recent advances in generative AI has led to the development of text-to-video models such as Video Diffusion Models~\cite{ho2022video}, Make-A-Video~\cite{singer2022make}, Phenaki~\cite{villegas2022phenaki}, CogVideo~\cite{hong2022cogvideo}, ImagenVideo~\cite{ho2022imagen}, PyoCo~\cite{ge2023preserve}, PVDM~\cite{yu2023video}, etc. These pretrained text-to-video models, along with pretrained text-to-image models, have been used for video generation, video editing and personalization (including motion/ action personalization) by a number of approaches such as Dreamix~\cite{molad2023dreamix}, Tune-A-Video~\cite{wu2023tune}, Pix2Video~\cite{ceylan2023pix2video}, Align Your Latents~\cite{blattmann2023align}, CCEdit~\cite{feng2023ccedit}, Fatezero~\cite{qi2023fatezero}, video-p2p~\cite{liu2023video}, Wang et. al.~\cite{wang2023zero},ConditionVideo ~\cite{peng2023conditionvideo}, infusion~\cite{khandelwal2023infusion},Zhang et. al.~\cite{zhang2023towards}. However, these methods are usually limited to performing style transfer, concept editing and single subject personalization in videos where a single subject is involved. Methods such as Esser et. al.~\cite{esser2023structure} are able to control the structure and content individually, but require a large amount of data to train on. The goal of our paper is to generate a multi-concept customized video, consistent with the input text description, using a limited number of data samples per image/ video concept ($1$ in this paper) and a pretrained text-to-video model.

\subsection{Personalization of generative models}

Image and video personalization using generative AI models has seen tremendous progress in the recent past for single concept personalization with the development of methods such as DreamBooth~\cite{ruiz2023dreambooth}, ControlNet~\cite{zhang2023adding}, Instructionpix2pix~\cite{brooks2023instructpix2pix}, Imagic~\cite{kawar2023imagic}, Null text inversion~\cite{mokady2023null}, Video ControlNet~\cite{hu2023videocontrolnet}, Attend-and-excite~\cite{chefer2023attend}, paint by example~\cite{yang2023paint}, Free-bloom~\cite{huang2023free}. Recent times have also seen multi-concept customization methods for text-to-image models such as Custom Diffusion~\cite{kumari2023multi}, FastComposer~\cite{xiao2023fastcomposer}, SVDiff~\cite{han2023svdiff}, Subject Diffusion~\cite{ma2023subject}. However, multi-concept customization methods developed for text-to-image models are not very effective~\cite{chen2023videodreamer} for video customization and cause artifacts, attribute binding problems, memory and data issues. In contrast, our method trains a single lightweight adapter jointly on all concepts and does not require any data regularization for inducing variance. 
VideoDreamer~\cite{chen2023videodreamer} is a multi-subject video customization method concurrent to our work. VideoDreamer leverages a pretrained text-to-image stable diffusion model to extend to custom video generation. Its use of a text-to-image model limits its application to cases where there is minimal subject-subject interaction or limited motion. Moreover, it cannot transfer or customize motion information. In contrast, our goal is to leverage pretrained text-to-video models to perform multi-concept customization, where the custom concept can be a subject or motion, or even background.

\section{Method}

\subsection{Problem Formulation}
We are given a pretrained text-to-video model $\mathcal{M}$ and image(s) or video(s) $c_{i}, i = 1...N$ corresponding to $N$ concepts with text descriptions $t_{c_{i}}, i = 1...N$, where $N$ is the number of concepts. Our goal is to generate a video $V_{gen}$ that contains the custom concepts, as described by a query text $t_{gen}$. We assume that the model is capable of auto-regressively generating future frames. 

%Oct25:MBZ: We should mention that we are assuming the model is capable of auto-regressively generating more frames. While this is not a very small assumption we can under play it by mentioning the models that support it. Divya - done

\subsection{Proposed Method}
We present an overview of our method in Figure~\ref{fig:overview}. As our first step, we incorporate the knowledge of the custom concepts $c_{i}, i = 1...N$ within the text-to-video model by finetuning the diffusion model. Our next step is to use the finetuned model to generate the custom video, corresponding to $t_{gen}$, by adding subjects one-by-one, in an autoregressive manner. Each step of the generation progressively adds various custom concepts and their corresponding interactions, as dictated by the query text $t_{gen}$. We now turn to describe our method in detail.

\subsubsection{Multi-concept finetuning}

\label{sec:train}

Given $c_{i}, i=1...N$, we fine-tune $\mathcal{M}$. Following~\cite{hu2021lora}, we do parameter efficient fine-tuning and specifically, adapter tuning, as is shown effective for generative image transformers~\cite{sohn2023visual}. Note that we jointly train a single adapter for all concepts instead of training an individual adapter for each concept as in~\cite{kumari2023multi}. 

Our next step is to use the finetuned text-to-video model to generate personalized videos with custom concepts, dictated by a query text $t_{gen}$. %While the finetuned T2V model is capable of generating customized videos containing a single custom concept, it is not very effective in generating videos with multiple custom concepts. This is due to two reasons: (i) the adapter model overfits on the concepts, which results in low variance; variance is important for the generation of interactions between concepts in the video (ii) the inability of the model to understand the interactions between various custom concepts due to generalization issues w.r.t. the dataset the model was pretrained on. 
Consider a \textit{Teapot} and a \textit{Tree}. The video manifold $\mathrm{M_{1}}$ of \textit{Teapots} contains videos of the \textit{Teapot} pouring tea, floating in the ocean, being washed, boiling tea and in a kitchen, dictated by text. Similarly, the video manifold $\mathrm{M_{2}}$ of \textit{Trees} contains videos of the tree swaying in the wind, the relational notion of under the tree, in a park, squirrels playing around the tree, cat climbing the tree and the tree being located in a dense forest, dictated by text. Thus, each of the manifolds contains what the model is familiar with in terms of generation. The video corresponding to "a \textit{Teapot} boiling tea under a \textit{Tree}" is contained in the (non-linear) intersection of the two manifolds $M_{1} \cap M_{2}$, generation of which is not straightforward. In the next section, we present a method to generate the multi-concept customized video.

\subsubsection{Causal generation, one subject at a time}
%Oct25:MBZ: maybe rename to something less generic such as "Causal generation, one subject at a time.": Done

We hypothesize that sequential and controlled traversal through the various manifolds (each understood by the model) towards the intersection space of the manifolds, while generating video frames sequentially, leads to the solution. Our solution is based on multi-step generation, each step incrementally generating what the model understands, while remembering what it has already generated. Thus, we begin at the manifold corresponding to one of the custom concepts to generate a few frames corresponding to the first concept by using an appropriate text description. Next, to generate video frames with multiple concepts, we condition on the generated video frames of the first concept and use text descriptions describing the other concepts to incrementally add them and their interactions. 

For instance, in the case of "a \textit{Teapot} boiling tea under a \textit{Tree}", we begin at the manifold of \textit{Trees} to generate a few frames of the \textit{Tree} using the text prompt "a \textit{Tree}". Next, move towards the manifold of \textit{Teapots} to generate a video of the \textit{Teapot} boiling tea under a \textit{Tree}. Consequently, to generate the next video frames of "a \textit{Teapot} boiling tea under a \textit{Tree}", we use the text prompt "a \textit{Teapot} boiling tea under a \textit{Tree}", conditioned on the previously generated frames of the \textit{Tree}. Thus, the first few frames of the video depict the \textit{Tree}, and the subsequent frames show the \textit{Teapot} boiling tea under the \textit{Tree}.

At each step of the sequential generation, the existence of a strong prior (in terms of video frames generated upto that stage), along with the variance arising from the pretrained model, alleviates bias related issues. Consequentially, at each step of the sequential generation, the model needs to be conditioned on an appropriate number of prior frames it previously generated and generate the subsequent frames in an autoregressive manner. If too few frames are used for the conditioning, the model cannot remember prior knowledge. On the other hand, if too many frames are used, the model may not be able to bring in new concepts due to bias issues. Moreover, it is important to keep the sequential generation controlled, in order to make sure that the generated video stays within the intersection manifold i.e. the number of frames generated for each concept is a key hyperparameter. In summary, causality is key to maintain prior context. % we leverage the strong causality in Phenaki to achieve the same. 

%In summary, adding complexity sequentially rather than at once makes it easier for the model to generate the multi-concept customized video. Causality, when leveraged appropriately, helps in generating the interactions between the various concepts in the video. Prior knowledge helps in decreasing the bias towards the single image corresponding to the concept, 
Algorithmically, let $F$ be the total number of frames that are being generated and $p_{i}, i=1...N$ be the prompts describing the various concepts and their interactions. Tokens corresponding to all $F$ frames are initialized to be empty. 
\begin{equation}
\text{for } k \text{ in range } (0, F): z_{k} \leftarrow \mathcal{M}(z_{k}) | (z_{k-1:k-m}, p). 
\end{equation}
As per this equation, the generation of each frame (except the first $m$ frames) is conditioned on the past $m$ frames. %Conditioning on $m$ frames, determined empirically, is consistent with the fact that Phenaki is trained to generate $11$ frames at a time and conditioning on half the frames achieves the right trade-off between the amount of prior knowledge required for the generation of future frames. 
The prompt $p$ is sequentially set to $p_{1} ... p_{N}$. Since the model can generate $11$ frames in one go, we set $p = p_{1}$ for $k<=11$. In the 2-concept customization case, we set $p = p_{2}$ for the rest of the frames.

\paragraph{Structuring the concepts.} Sequential generation requires the various concepts to be structured in an appropriate manner -- it is important to carefully design $p_{i}, i = 1...N$. We propose to structure the concepts in a top-down manner - consistent with the natural representation of a scene. For example,
\begin{itemize}[nosep]
    \item Custom subject and custom background - first generate the background. Next, condition on the frames containing the custom background to add the custom subject performing action. For eg. given concepts `a \textcolor{RowColorCode}{B1*} futuristic restaurant' and `a \textcolor{RowColorCode}{C2@} cat', and $t_{gen} =$ `a \textcolor{RowColorCode}{C2@} cat eating noodles in a \textcolor{RowColorCode}{B1*} futuristic restaurant', we set
    \begin{itemize}[nosep]
        \item $p_{1} =$ `a \textcolor{RowColorCode}{B1*} futuristic restaurant'
        \item $p_{2} =$ `a \textcolor{RowColorCode}{C2@} cat eating noodles in the futuristic restaurant'.
    \end{itemize}
    \item Custom subject and custom action (from a video) - first generate a random (related subject) performing the custom action. Next, steer the model towards the manifold of the custom subject performing action. For eg. given concepts `a person playing tennis in a tennis court' and `a \textcolor{RowColorCode}{C2@} cute cat', and $t_{gen} =$ `a \textcolor{RowColorCode}{C2@} cute cat playing tennis in a tennis court', we set
    \begin{itemize}[nosep]
        \item $p_{1} =$ `a person playing tennis in a tennis court', 
        \item $p_{2} =$ `a \textcolor{RowColorCode}{C2@} cat playing tennis in a tennis court'.
    \end{itemize}
    \item Custom subject 1 and custom subject 2 - generate one of the custom subjects. Condition on this to generate the other custom subject interacting with it. For eg. given concepts `a \textcolor{RowColorCode}{B1*} brown teapot' and `a \textcolor{RowColorCode}{C2@} tree', and $t_{gen} =$ `a \textcolor{RowColorCode}{B1*} brown teapot boiling tea under a \textcolor{RowColorCode}{C2@} tree', we set \begin{itemize}
        \item $p_{1} =$ `a \textcolor{RowColorCode}{C2@} tree',
        \item $p_{2} =$ `a \textcolor{RowColorCode}{B1*} brown teapot boiling tea under a \textcolor{RowColorCode}{C2@} tree'.
    \end{itemize}
\end{itemize}

\section{Experiments}

\begin{figure}
    \centering
    \includegraphics[scale=0.3]{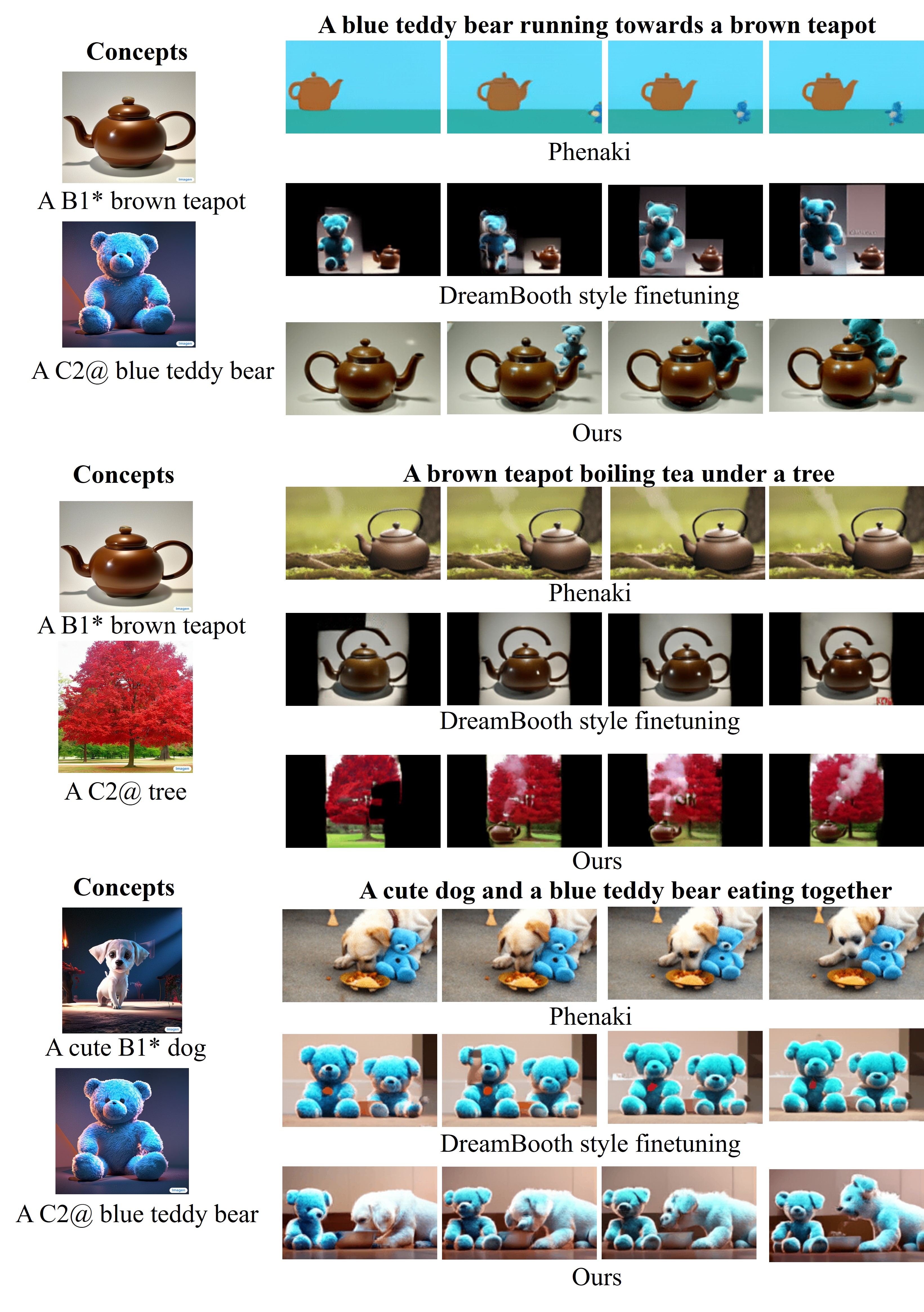}
    \caption{\textbf{Two subject customization.} While Phenaki contains some prior knowledge of the interactions between different kinds of subjects, the finetuned model is unable to generate the interactions between the custom subjects. Through controlled and sequential autoregressive generation of concepts and their interactions, our method is able to generate customized videos with two subjects interacting with each other.}
    \label{fig:results1}
\end{figure}
\begin{figure}
    \centering
    \includegraphics[scale=0.3]{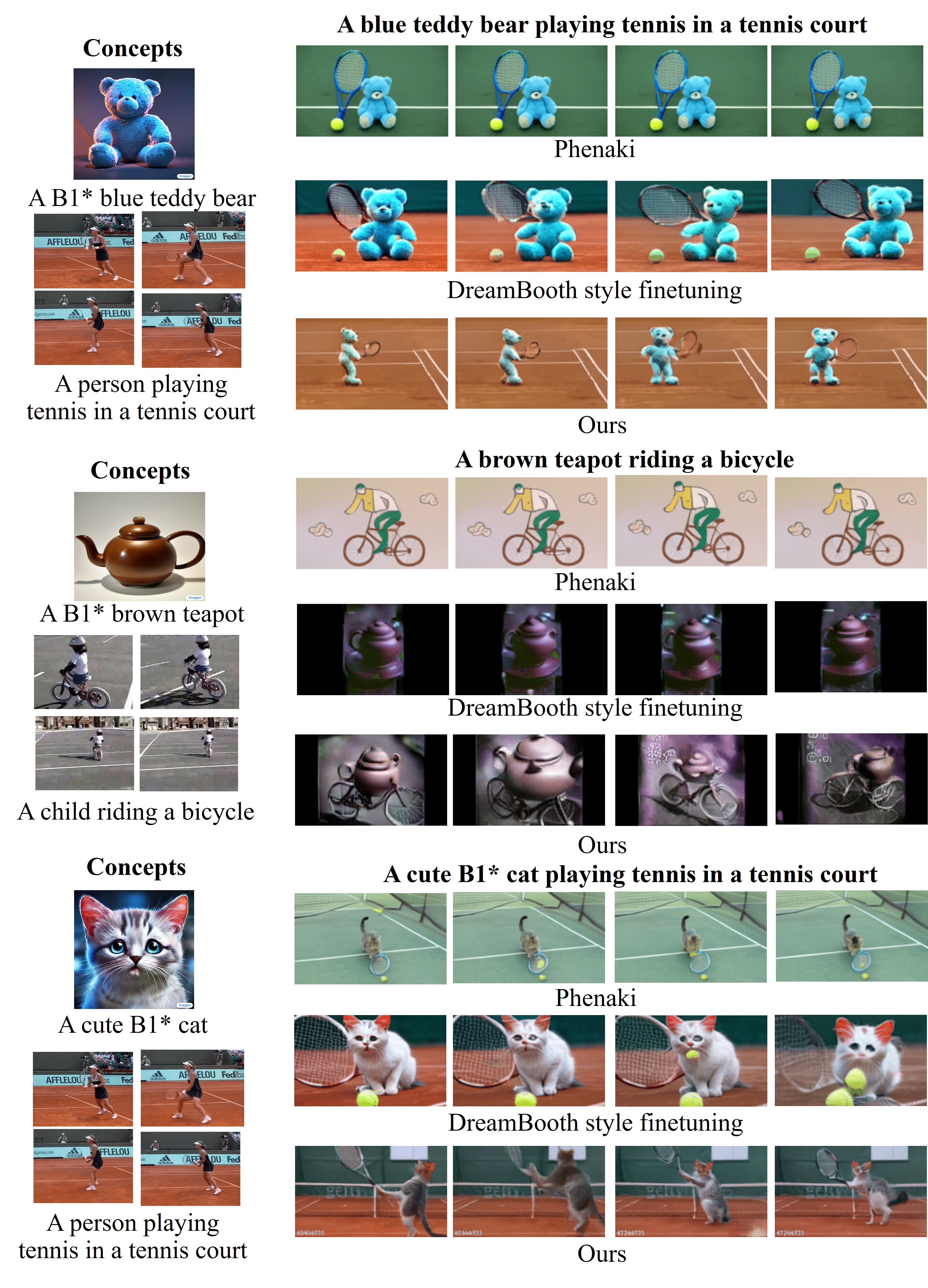}
    \caption{\textbf{Subject Action customization.} Phenaki, due to lack of prior knowledge, is unable to generate certain actions. The finetuned model is able to generate the custom subject, but not the precise action, even after finetuning. Generating the motion corresponding to the action first, followed by generating the custom subject performing action (conditioned on the motion) enables the generation of the custom subject performing the custom action. Thus, we are not only able to `teach' the action to the model, but are also able to customize the subject performing the action.}
    \label{fig:results2}
\end{figure}
\begin{figure}
    \centering
    \includegraphics[scale=0.3]{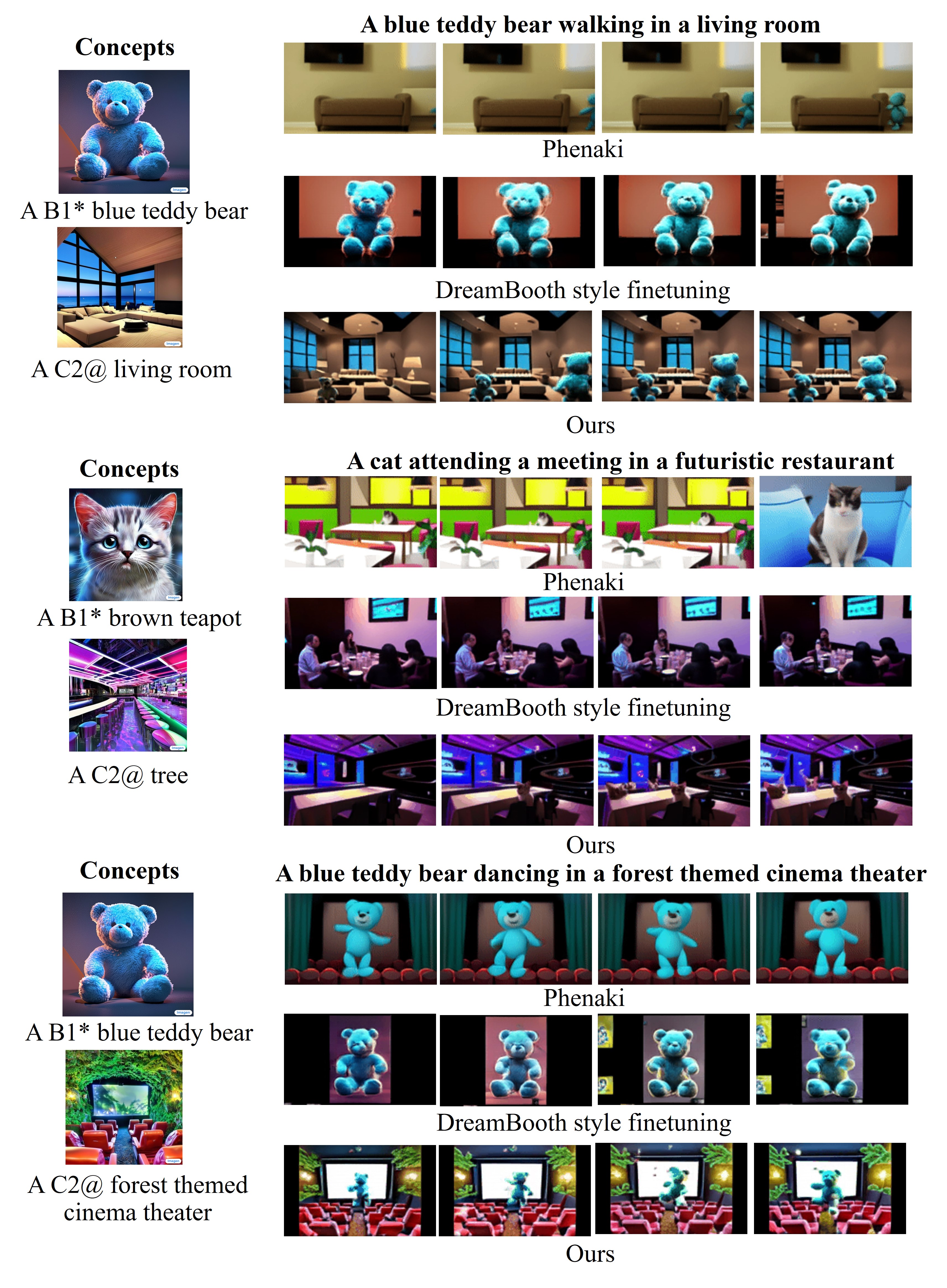}
    \caption{\textbf{Subject BG customization.} While subject-background customization is relatively easy for the Phenaki model, the finetuned model is unable to emphasize on the background due to bias issues. This is resolved by our model, which first generates the background and then generates video frames of the custom subject performing action, conditioned on the background.}
    \label{fig:results3}
\end{figure}

\subsection{Setup}

Our proposed method for multi-concept video customization is built on autoregressive generation. Consequentially, we use Phenaki~\cite{villegas2022phenaki}, a text-to-video model with autoregressive capabilities, as our backbone architecture. Phenaki learns video representations in a causal manner, compressing the video into a small representation of discrete tokens using a spatial transformer and a causal transformer. A MaskGiT transformer model~\cite{chang2022maskgit} is then trained to reconstruct the video tokens, conditioned on a text prompt. A bidirectional text transformer is used to compute video tokens from text. At inference, starting from empty tokens and conditioned on the input text, video tokens are generated, which are then detokenized to generate a video. At each step of the inference, Phenaki can freeze video tokens corresponding to past frames and generate video tokens corresponding to future frames in an autoregressive manner. Classifier-free guidance~\cite{ho2022classifier} controls the alignment between the text and the generated video. Phenaki is pretrained on a large-scale text-image and text-video dataset, which we use for multi-concept video customization. We use the low-resolution Phenaki model in all experiments. 

%Oct25:MBZ: this needs a ref since it's a big general claim. I also don't think it's necessary :) maybe we can restructure below by saying the a "custom subject" can take multiple forms e.g. background, visuals of a character and the action taken. We experiment with various mix and match of such objects to the point that the quality of the model allows: Done 

We finetune the adapter weights jointly on all image concepts, followed by video concepts, for 1000 steps with a learning rate of $1e-4$ using the Masked Visual Token Modeling (MVTM) loss\cite{chang2022maskgit}, along with classifier-free guidance. The adapter has a hidden size of $2$. Our generated videos have a resolution of $160 \times 96$. At inference time, to condition on prior frames, we set $m = 5$. Conditioning on $5$ frames, determined empirically, is consistent with the fact that Phenaki is trained to generate $11$ frames at a time and conditioning on half the frames achieves the right trade-off between the amount of prior knowledge required for the generation of future frames. For each example, we present comparisons of our method with two baselines: 
\begin{itemize}
    \item DreamBooth style finetuning: In this case, the model is finetuned as described in Section~\ref{sec:train} and inferred using $t_{gen}$. This is inspired by the DreamBooth~\cite{ruiz2023dreambooth} model with adapter~\cite{hu2021lora} layers, extended to Phenaki video customization. This is our closest baseline for multi-concept video customization. 
    \item Phenaki: This baseline corresponding to Phenaki as it was trained with no additional fine-tuning.
\end{itemize} 
We use Imagen~\cite{ho2022imagen} to generate the image for each custom concept, concept-specific videos are from UCF-101~\cite{soomro2012ucf101}. Custom concepts can take multiple forms, e.g. background, visuals of a character or subject and a custom performed action. We experiment with various types of customization such as subject-subject, subject-action, subject-background. The design of the prompts was a purely creative task, we designed close to 100 prompts and attempted to keep it as diverse as possible. Our data set-up is as follows: 
\begin{itemize}
    \item Subject-subject customization: We use 14 examples corresponding to the 3 examples shown in the paper, 9 in the supplementary and `a cute dog and a blue teddy bear in a running race', `a cute dog washing a blue plate'.
    \item Subject-action customization: We use 4 subjects: a small cute fox, a small cute dog, a small cute cat and a blue teddy bear and 8 actions: playing violin, playing hula hoop, playing drums, playing tennis, crawling, doing pushups, riding bicycle, doing TaiChi. This results in a total of $32$ videos.
    \item Subject-background customization: We use 4 subjects: a small cute fox, a small cute dog, a small cute cat and a blue teddy bear, 4 backgrounds: living room, futuristic restaurant, forest themed cinema theater, ocean with seaplants and fishes. For each background, we generate 2 actions: (dancing, walking), (eating noodles, attending a meeting), (dancing, clapping hands), (playing with fishes, swimming) in order. This results in a total of $32$ videos.
\end{itemize}

\subsection{Qualitative Results}
\subsubsection{Two-concept customization}

\paragraph{Two-subject customization (Figure~\ref{fig:results1}):} In the first case, while Phenaki is able to generate the video of a generic teddy bear running towards a generic teapot, the teddy bear is much smaller than the teapot. The finetuned model generates both subject in the video but is unable to generate the correct interaction (of the teddy bear running \textit{towards} the teapot). Our method structures the generation as $p_{0} = $ `a \textcolor{RowColorCode}{B1*} brown teapot' and $p_{1} = $ `a \textcolor{RowColorCode}{C2@} blue teddy bear running towards the \textcolor{RowColorCode}{B1*} brown teapot' to generate a video of the \textcolor{RowColorCode}{C2@} blue teddy bear running towards the \textcolor{RowColorCode}{B1*} brown teapot; the static object (teapot) is generated first followed by the moving teddy bear. In the second example, Phenaki is able to generate the video with generic subjects. The finetuned model has a bias towards the teapot and fails to generate the tree. Our method structures the generation as $p_{0} = $ `a \textcolor{RowColorCode}{C2@} tree' and $p_{1} = $ `a \textcolor{RowColorCode}{B1*} brown teapot boiling tea under a \textcolor{RowColorCode}{C2@} tree' to generate a customized video of the teapot boiling tea under the tree; the background, static subject (tree) is generated first followed by the teapot boiling tea. In the third example, Phenaki generates a toy teddy bear, that is not eating. The finetuned model mixes the attributes of the concepts. Our method structures the generation as $p_{0} = $ `a cute \textcolor{RowColorCode}{B1*} dog eating together with the \textcolor{RowColorCode}{C2@} blue teddy bear' and $p_{1} = $ `a \textcolor{RowColorCode}{C2@} blue teddy bear eating together with a cute \textcolor{RowColorCode}{B1*} dog' to generate the dog and teddy bear eating together; the emphasis is on the dog first, followed by the teddy bear. In summary, controlled and sequential generation of the custom concepts and their interactions is effective in generating a multi-subject customized video. 

\paragraph{Subject-action customization (Figure~\ref{fig:results2}):} We first generate the motion depicting the action, followed by the subject performing the action conditioned on the motion frames. Hence, $p_{0}, p_{1}$ take the general structure of $p_{0} = $`a random subject performing the action', $p_{1} = $`custom subject performing the action'. In the first case, the pretrained model Phenaki does not understand what a teddy bear playing tennis looks like and is unable to generate the video. In the result of the fine-tuned model, there is some movement in terms of the teddy bear holding the tennis bat but the video does not depict a teddy bear playing tennis. Our method is able to generate a video of the custom teddy bear playing tennis. Thus, our method is not only able to teach the model the action `playing tennis', but is also able to effectively condition on the motion to customize the subject performing action. Similarly, in the third example, neither Phenaki nor the finetuned model are able to generate a cat playing tennis. Our method is able to generate a video of a custom cat playing tennis, thus teaching the model the action, as well as customizing the subject. In the second example, again, the Phenaki model does not know what a teapot riding a bicycle is. The finetuned model is able to generate some movement of the teapot but the generated video isn't really of a teapot riding a bicycle. 

\paragraph{Subject-background customization (Figure~\ref{fig:results3}):} Our next task is to customize the subject and the background (and generate a random action). We first generate the custom background, followed by the subject performing a generic action in the background. Hence, $p_{0}, p_{1}$ take the general structure of $p_{0} = $`custom background', $p_{1} = $`custom subject performing action in the background'. In the first example, while Phenaki understands what a teddy bear walking in a living room is, when finetuned, due to bias issues, the finetuned model is unable to generate a custom teddy bear walking in the custom living room. Our method first generates the custom living room, conditioned on which, the custom teddy bear walking in the living room is generated. Similarly, in the second and third examples, while Phenaki understands the interactions between the subject and the background, after finetuning, due to bias issues, the model is unable to generate the background well. Our method is able to generate videos of custom cats attending meetings in futuristic restaurants, and a custom teddy bear dancing in a forest themed cinema theater. 
\begin{figure}
    \centering
    \includegraphics[scale=0.3]{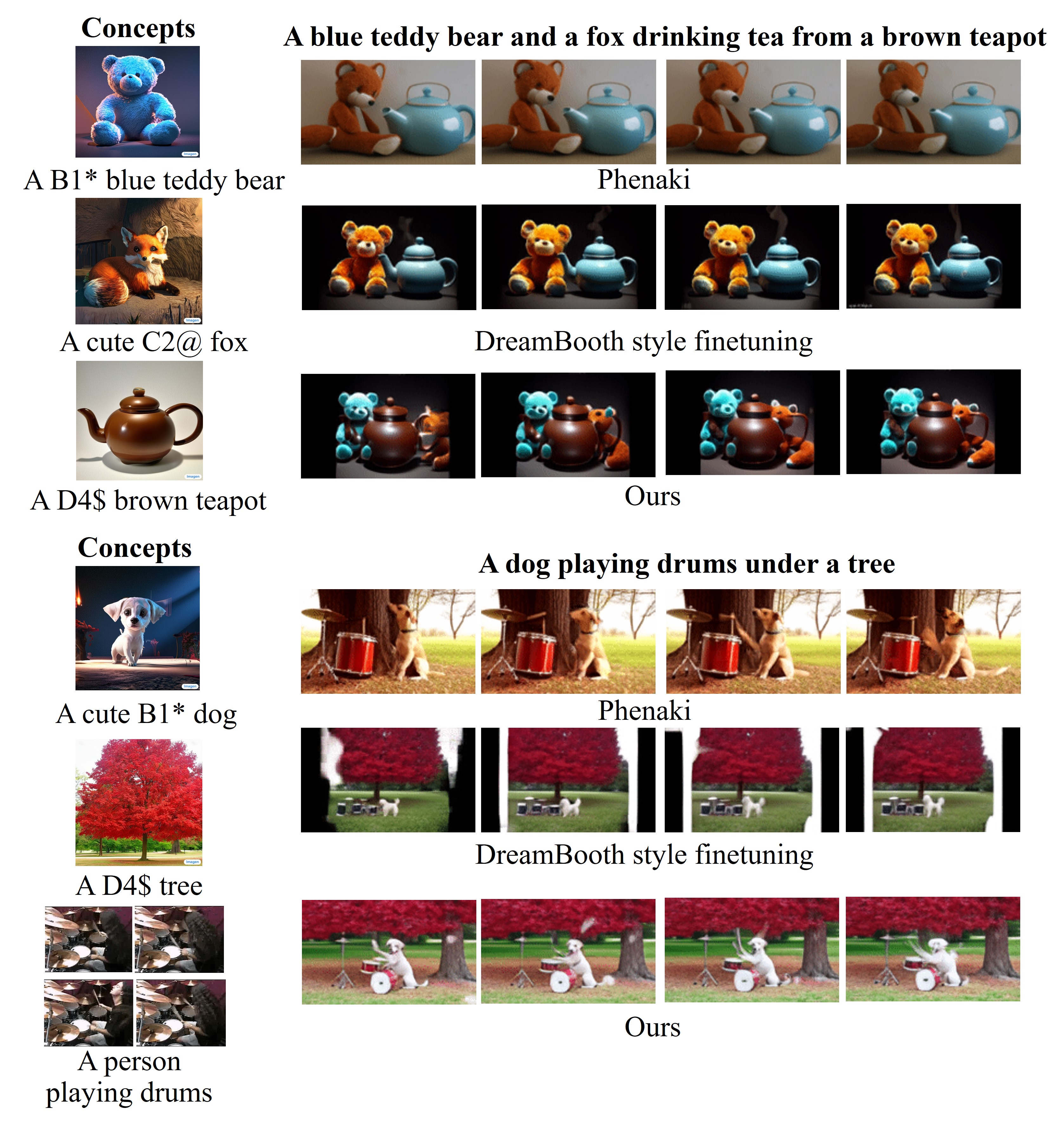}
    \caption{\textbf{Three-concept customization.} In the first case, the baseline models mix the attributes of the various concepts. Our method, by sequential generation of concepts and interactions, is able to generate the complex video with all desired concepts and interactions in the scene. Similarly, in the second example, while the finetuned model is able to bring in all concepts into the video, it is unable to generate the interaction of playing drums. Conditioning on motion information helps our method do so.}
    \label{fig:3concept}
\end{figure}
\begin{figure}
    \centering
    \includegraphics[scale=0.3]{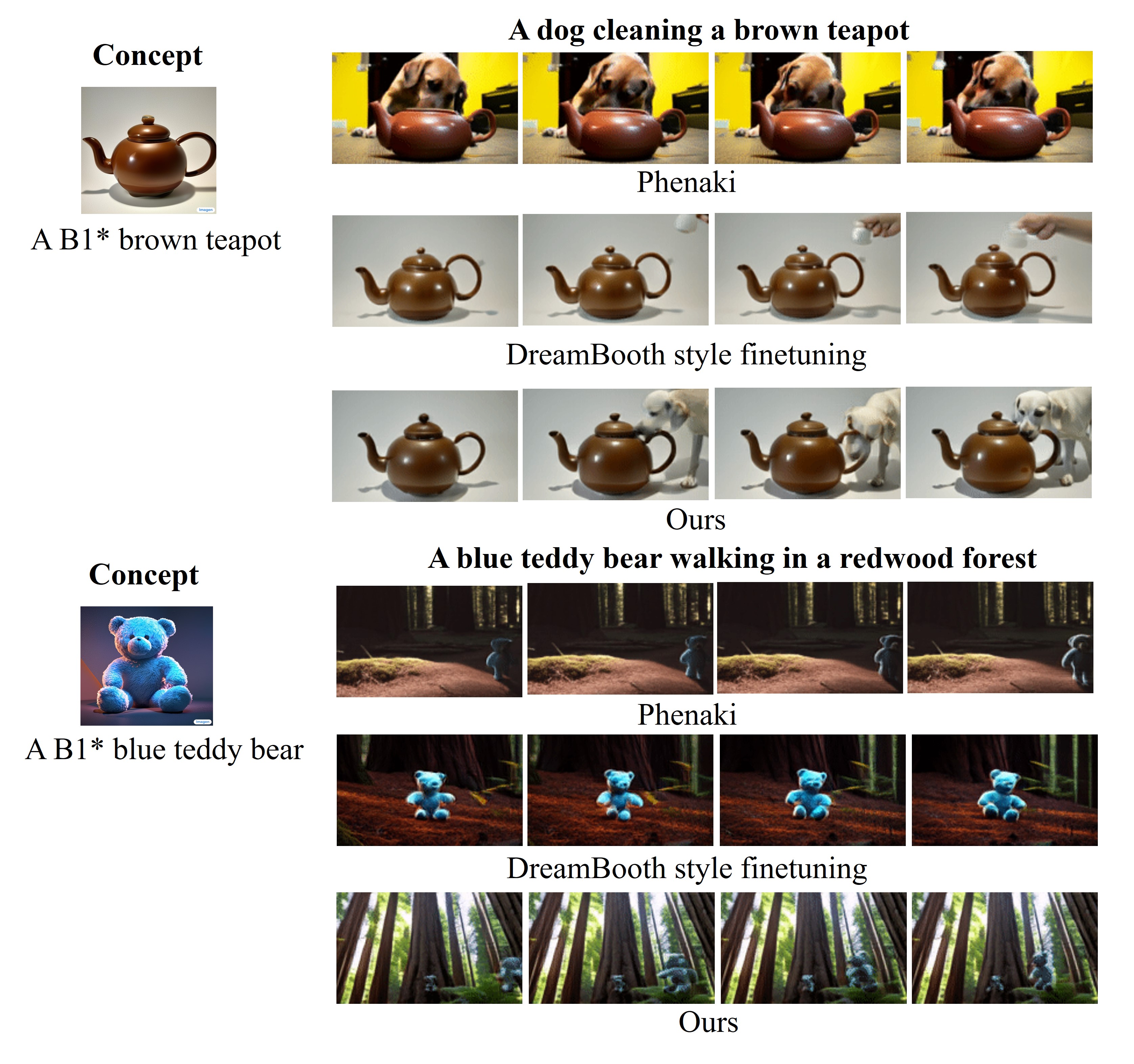}
    \caption{\textbf{Single concept customization and compositionality.} Causality can be a useful tool for generating videos where compositionality is desired, an interesting direction for future work.}
    \label{fig:singleconceptcompositionality}
\end{figure}

\begin{figure*}
    \centering
    \includegraphics[scale=0.6]{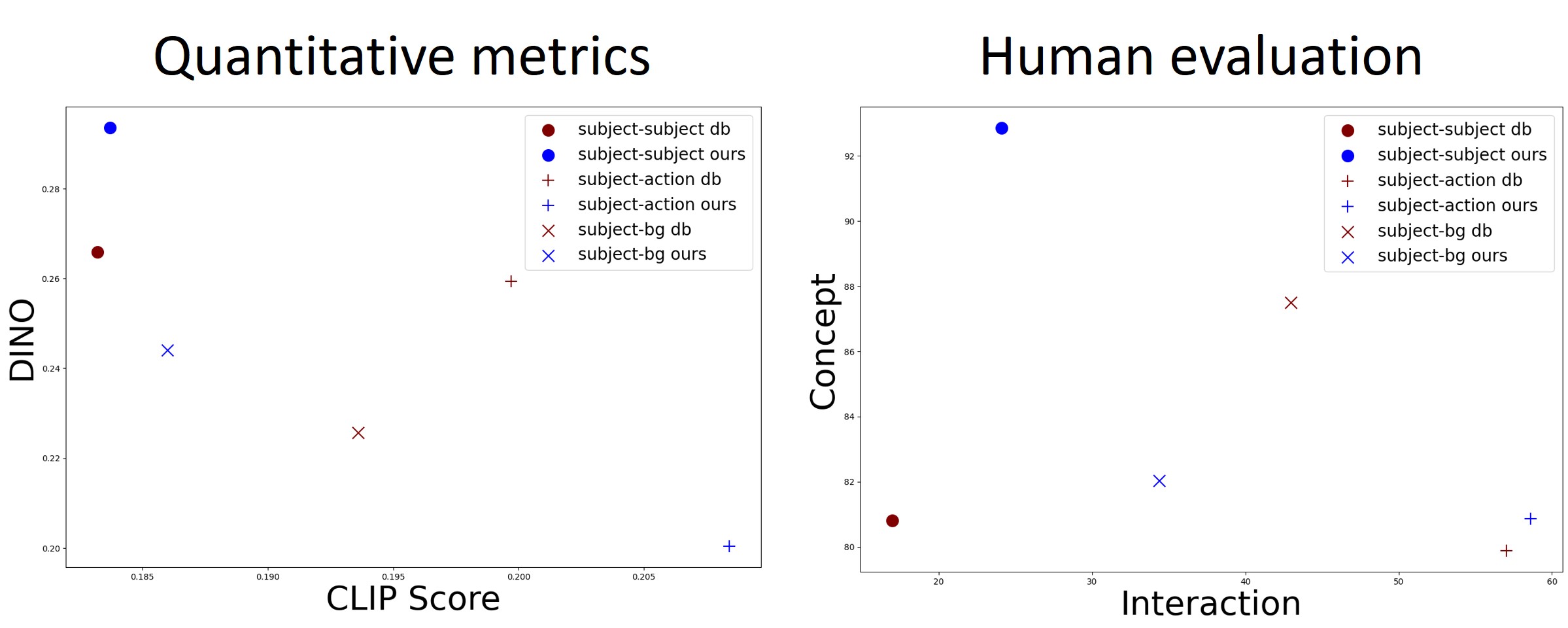}
    \caption{In the first figure, we show the videoCLIP and DINO scores for our method and finetuning baseline. In the second figure, we show results of the human evaluation. We present the analysis in Section~\ref{sec:quant}.}
    \label{fig:quantitative}
\end{figure*}

\subsubsection{Three-concept customization}

We show generated videos with 3 custom concepts in Figure~\ref{fig:3concept}. In the first example, we have three concepts: a teddy bear, a fox and a teapot and we want to generate a video of the two animals drinking tea together from the teapot. Phenaki is unable to generate the video (even without customization) - we see that the attributes of the three concepts are mixed up. The model generates a blue teapot instead of a brown teapot and a hybrid of the teddy bear and the fox. In the result of the finetuned model, a teddy bear in the color of the fox is generated, and the teapot is in the color of the teddy bear. There is no fox in the scene. By bringing in the concepts and the interactions systematically, our method is able to generate the custom blue teddy bear drinking tea from the custom brown teapot with the custom fox. In the second example, we want to generate a dog playing drums under a tree. The dog, the tree and the action of playing drums are customized. Phenaki understands the scene and is able to generate a generic video. The finetuned model brings in all concepts into the scene but is unable to generate the action of `playing drums'. Our method, by conditioning on the motion tokens (of playing drums), is able to generate a video of the custom dog playing drums under the custom tree. 

\subsubsection{Compositional single-concept customization}

While the focus of our method is on multi-subject customization, it is also useful for single-concept customization where the model does not understand compositionality. In such cases naive finetuning may not work due to bias issues and leveraging causality to generate compositional scenes can lead to better results. We show two cases in Figure~\ref{fig:singleconceptcompositionality}. These results provide a direction for future work on using causality to generate compositional videos.

%Single concept customization results also - DreaMix 

% \paragraph{More results.} Please refer to the supplementary material for more results.

\subsection{Quantitative results}

\label{sec:quant}

We quantitatively evaluate our method using the Video-CLIP score~\cite{xu2021videoclip} and the self-supervised similarity score, DINO~\cite{caron2021emerging}. The former computes the alignment with text, signifying if the concepts and interaction described by the text are present in the generated video. To compute the DINO score, we compute the average over the individual DINO score w.r.t. each concept in the video, also averaged over all frames of the video. Blau et. al.~\cite{blau2018perception} state that generative models face difficulty in trading off perception-distortion. Similarly, we argue that videoCLIP-DINO (or text describing concept-concept interaction vs fidelity w.r.t. each concept) is a difficult trade-off to solve as it relates to the bias-variance trade-off. 

We also analyze our results with a human evaluation study. For each video, we ask participants three questions: (i) Is concept 1 present in the video? (ii) Is concept 2 present in the video? (iii) Is concept 1 interacting with concept 2 as described by the text? We report the percentage of affirmative responses for each question and for each method. 

We show the results in Figure~\ref{fig:quantitative}. We use 14 examples for subject-subject customization, 32 examples for subject-motion customization and 32 examples for subject-background customization. To compute the DINO and videoCLIP scores, for each example, we generate 8 results and choose the generated video with the best videoCLIP score as the final result. For the human evaluation, we show all 8 results (for each example) to human raters and compute the average.

\paragraph{Subject-subject customization.} We show improved DINO score, there is a small improvement in the video-CLIP score as well. Human raters also rate our method much higher than the baseline in terms of both concept interaction, as well as alignment of the characteristics of the concept with the input images. This indicates our method's ability to better generate multiple custom subjects interacting with each other.

\paragraph{Subject-action customization.} We show improved videoCLIP score indicating our methods' ability to generate the action better. The DINO score of our method is lower than that of the baseline. The baseline method is unable to generate good actions and in many cases reproduces the input concept image with mild movement, hence, its resemblance to the input concept image is better leading to a higher DINO score. Human raters rate our method better in terms of both interaction as well as concept fidelity.

\paragraph{Subject-background customization.} We show improved DINO score, indicating our methods' ability to generate the subject as well as the background with high fidelity. The video-CLIP score of our method is lower than that of the baseline. This is because the baseline method generates a video that focuses on the subject performing the action. This also implies higher resemblance to the input subject image or a higher individual DINO score w.r.t. the subject. We believe this also leads to human raters preferring the baseline method more than our method. However, the background is not very well depicted. The individual DINO score for background for our method is $0.3496$, while that of the baseline is $0.1896$. Our method focuses on generating the entire background depicted in the input concept image, along with the subject performing action; thus indicating higher consistency with both input concepts while generating the interaction. 

\paragraph{Other comparisons.} Custom diffusion~\cite{kumari2023multi} requires us to mine images from the original training dataset
(on which Phenaki was trained) for regularization, which was unavailable. Other multi-concept
customization approaches, with the exception of
DreamBooth, were simply not feasible to expand to video customization using Phenaki.

\section{Limitations and Future Work}

Our method has a few limitations: (i) extension beyond three concepts is non-trivial because the model may not have the ability of remember what it generated in the first step, and structuring the concepts/ interactions is complex. (ii) Controlling the interactions through text is not easy, it may be useful to have more control signals that enable the model to generate interactions respecting the 3D world (such as contact points between subject and ground, relative size between objects). (iii) We use the low resolution model of Phenaki to generate the results. This was the only autoregressive model we had access to at the time of submission of the paper. The development of better video foundation models and superresolution can enable the generation of better results. (iv) The quality of the generated videos is highly dependent on the base model, with the development of richer models that understand temporal coherence and 3D relations in the world, the artifacts in the generated videos can be removed. Furthermore, the recent development of autoregressive diffusion transformer models for video generation can enable the application of our multi-concept video customization method. (v) Our method takes in a single image per subject to generate personalized videos. This leads to low variance, and the model hallucinates a lot of details corresponding to the subject. This causes low subject fidelity in some generated videos, a direction for future work.
%(iii) Relative sizes are not always respected. For instance, the teapot should be smaller than the teddy bear, the dancing fox should be much smaller than the overall dimensions of the living room. 
Future work can focus on overcoming these limitations, in addition to further improving the text alignment and fidelity w.r.t. input concepts. Another direction for future work is to automate the structuring of the prompts using LLMs and simplify hyperparameter tuning related to finding the number of future frames ( and conditioning on prior frames) to be generated using each prompt. Also, incorporating causality within non-causal text-to-video models such as video diffusion models to use them for multi-concept video customization is an avenue to explore. This paper provides a direction for the importance of causality in multi-concept customization and can also be extended to long video generation.

%\begin{figure}
%    \centering
%    \includegraphics[scale=0.27]{Figures/dreamboothlora_singleconcept1.jpg}
%    \caption{Customization methods such as DreamBooth LoRA, adapted to video customization, are able to successfully perform single-concept personalization such as subject personalization (`a blue teddy bear walks into the kitchen and sits on the dining table') and motion customization (`a teddy bear playing tennis in a tennis court').}
%    \label{fig:dreamboothlora_singleconcept}
%\end{figure}

%\begin{figure}
%    \centering
%    \includegraphics[scale=0.27]{Figures/dreambooth_multiconceptdrawbacks.jpg}
%    \caption{Customization methods such as DreamBooth LoRA~\cite{ruiz2023dreambooth} are not able to perform multi-concept customization due to the inability to generate interactions between different custom concepts caused by generalization issues w.r.t. the dataset the pretrained model was trained on, bias issues w.r.t. the image(s) or video(s) corresponding to the custom concepts, and mix-up of attributes. }
%    \label{fig:dreambooth_multiconceptdrawbacks}
%\end{figure}

%\begin{figure}
%    \centering
%    \includegraphics[scale=0.25]{Figures/supp4.jpg}
%    \caption{Additional results on three-concept customization.}
%    \label{fig:supp4}
%\end{figure}

\begin{figure*}
    \centering
    \includegraphics[scale=0.4]{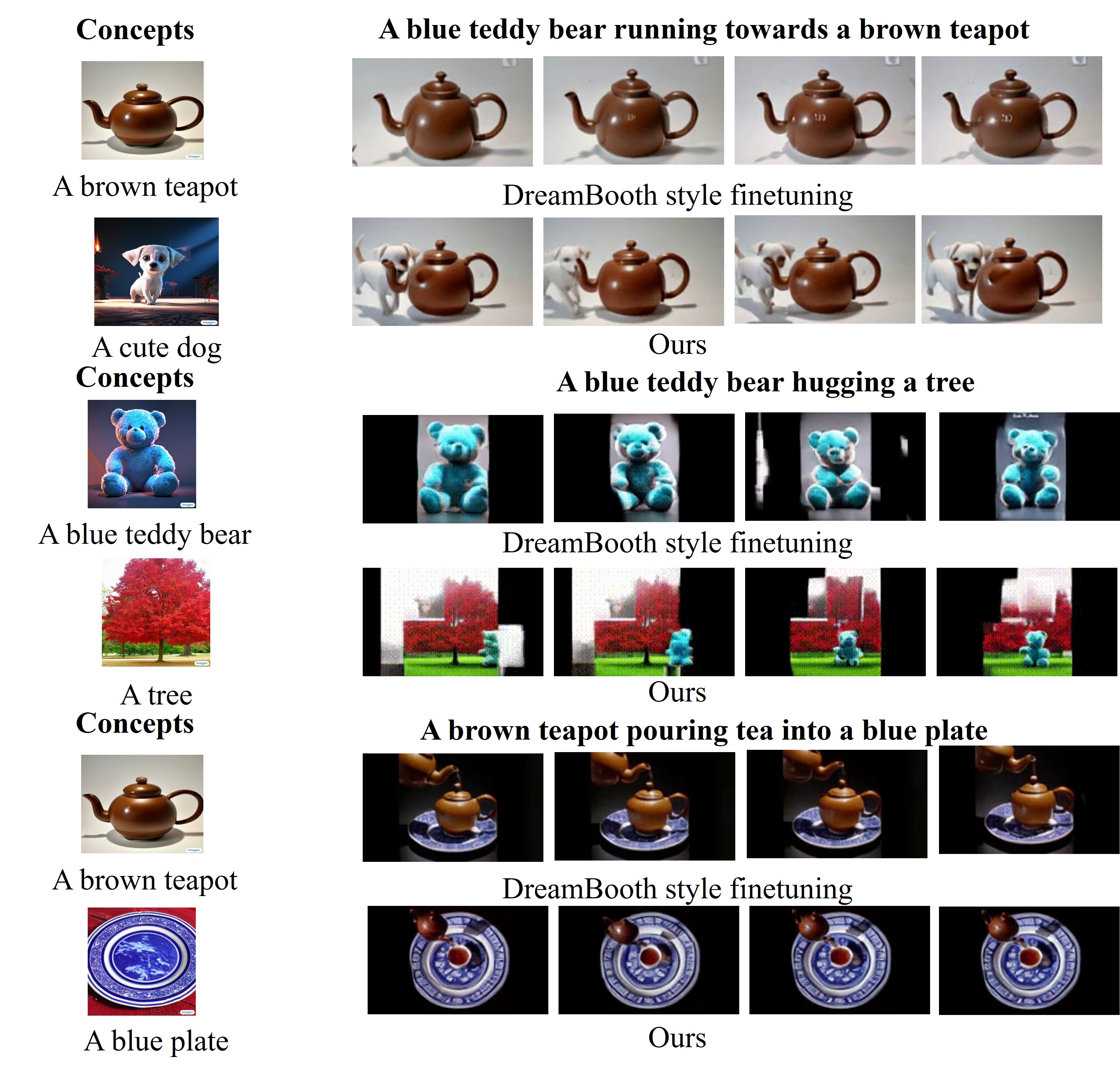}
    \caption{Additional results for subject-subject customization.}
    \label{fig:supp1}
\end{figure*}

\begin{figure*}
    \centering
    \includegraphics[scale=0.4]{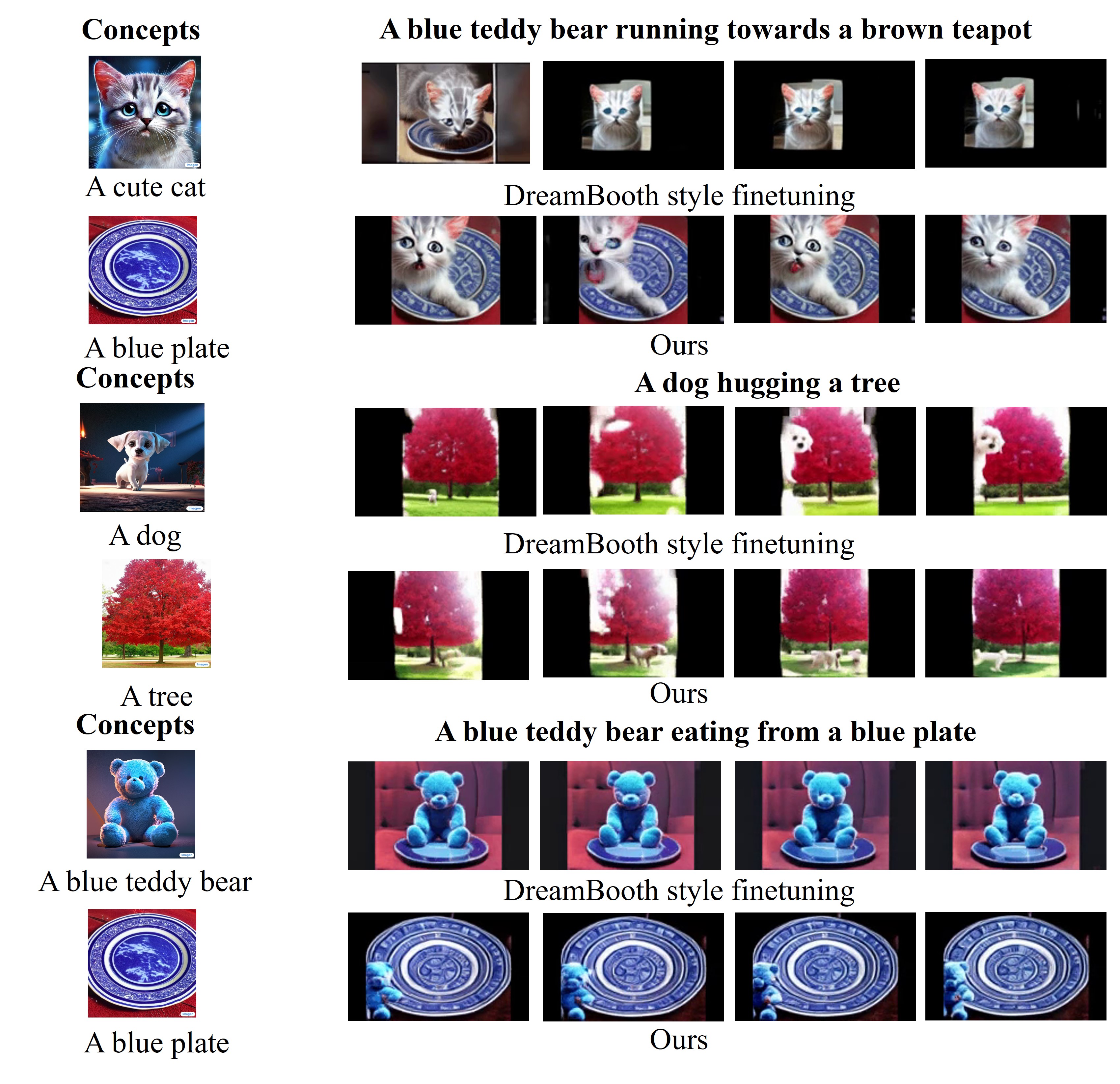}
    \caption{Additional results for subject-subject customization.}
    \label{fig:supp2}
\end{figure*}

\begin{figure*}
    \centering
    \includegraphics[scale=0.4]{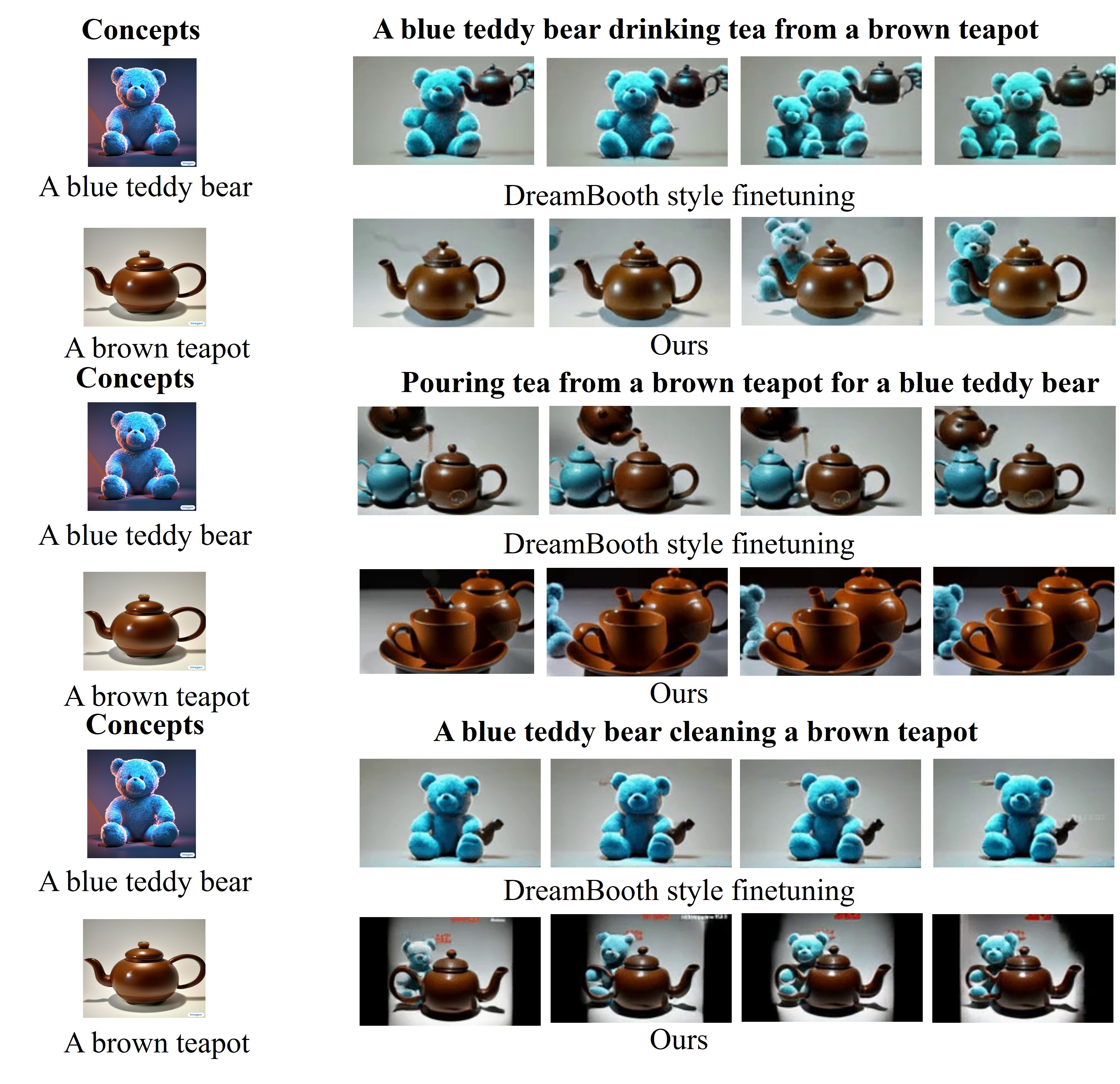}
    \caption{Additional results for subject-subject customization.}
    \label{fig:supp3}
\end{figure*}

\begin{figure*}
    \centering
    \includegraphics[scale=0.5]{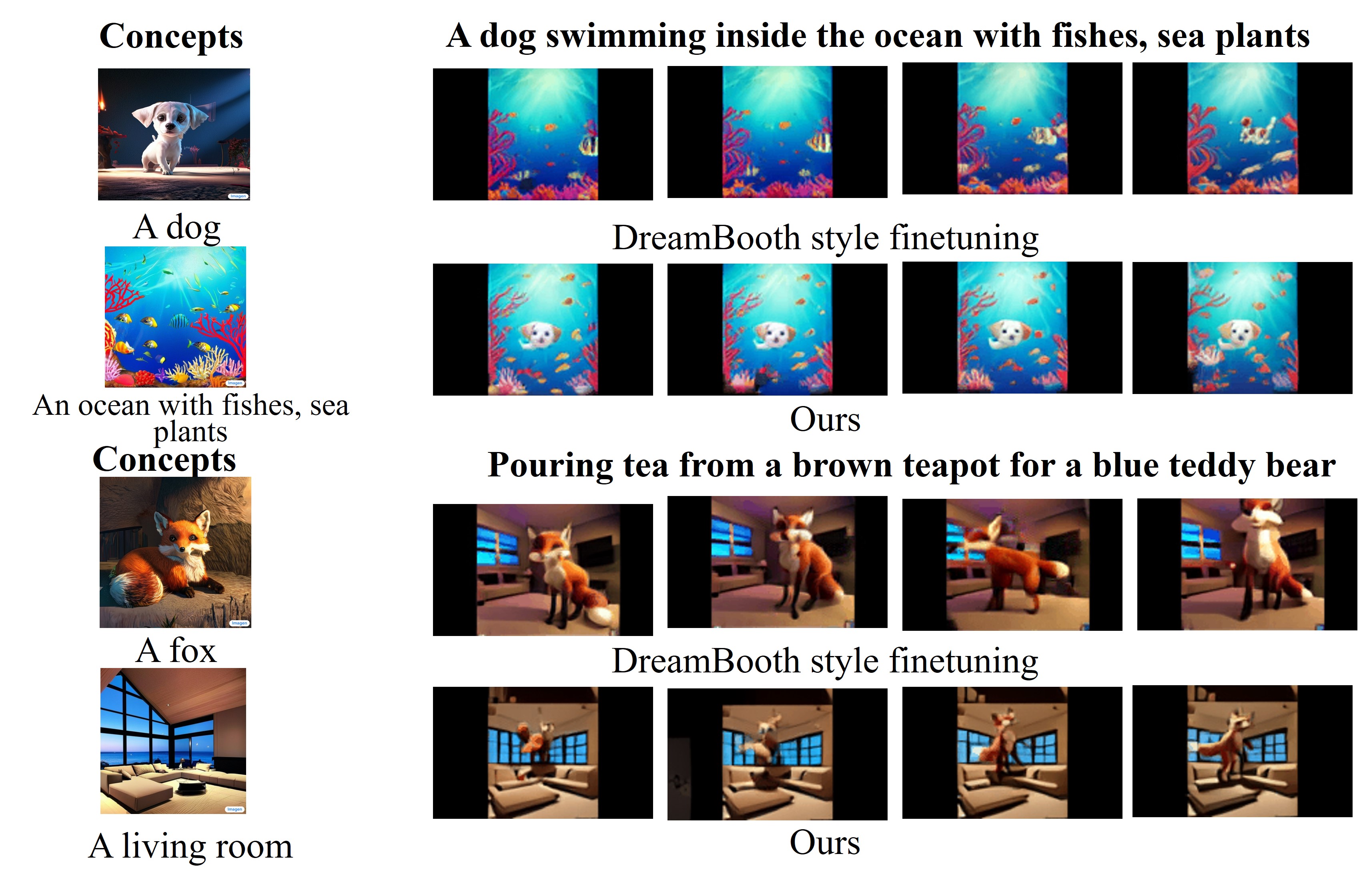}
    \caption{Additional results for subject-background customization.}
    \label{fig:supp5}
\end{figure*}

\begin{figure*}
    \centering
    \includegraphics[scale=0.5]{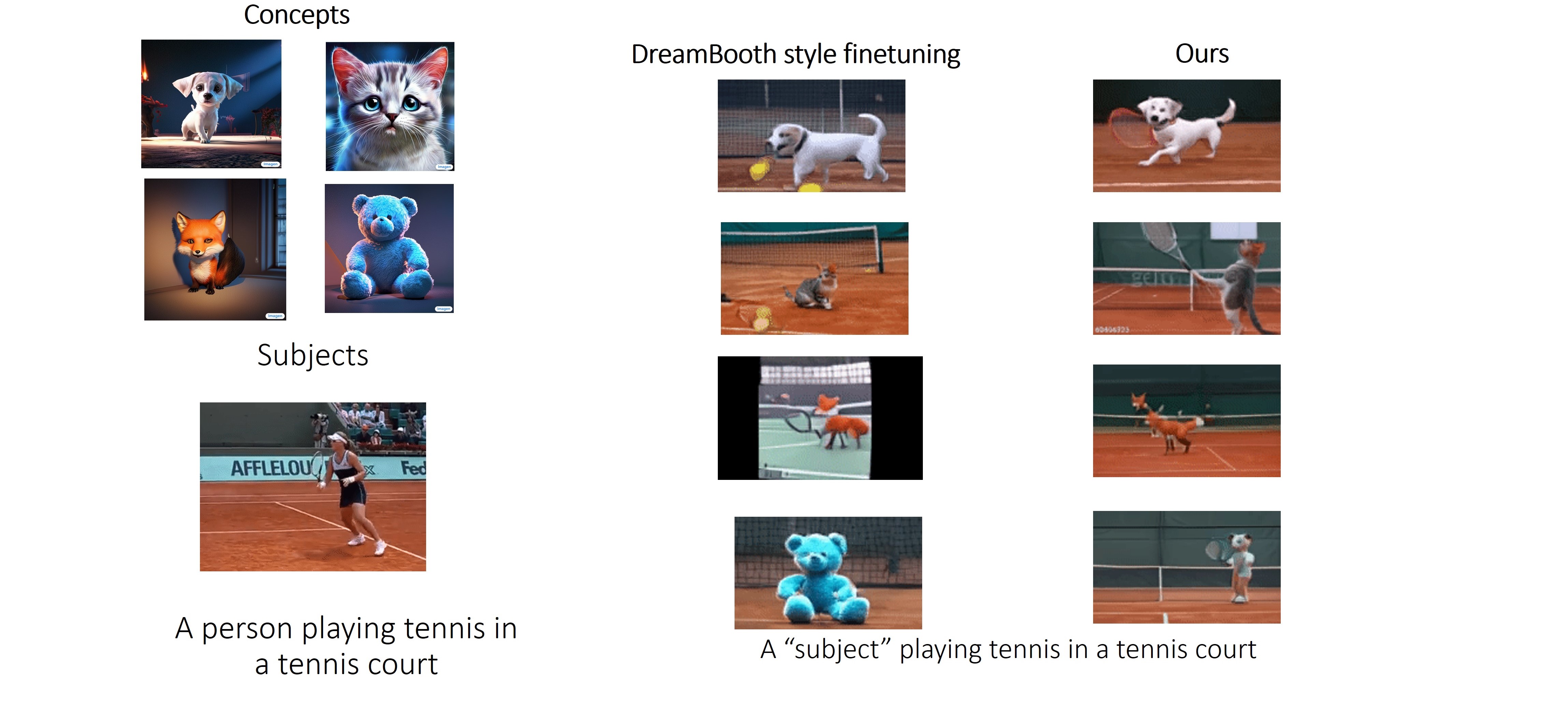}
    \caption{Additional results for subject-action customization. }
    \label{fig:supp6}
\end{figure*}

\begin{figure*}
    \centering
    \includegraphics[scale=0.5]{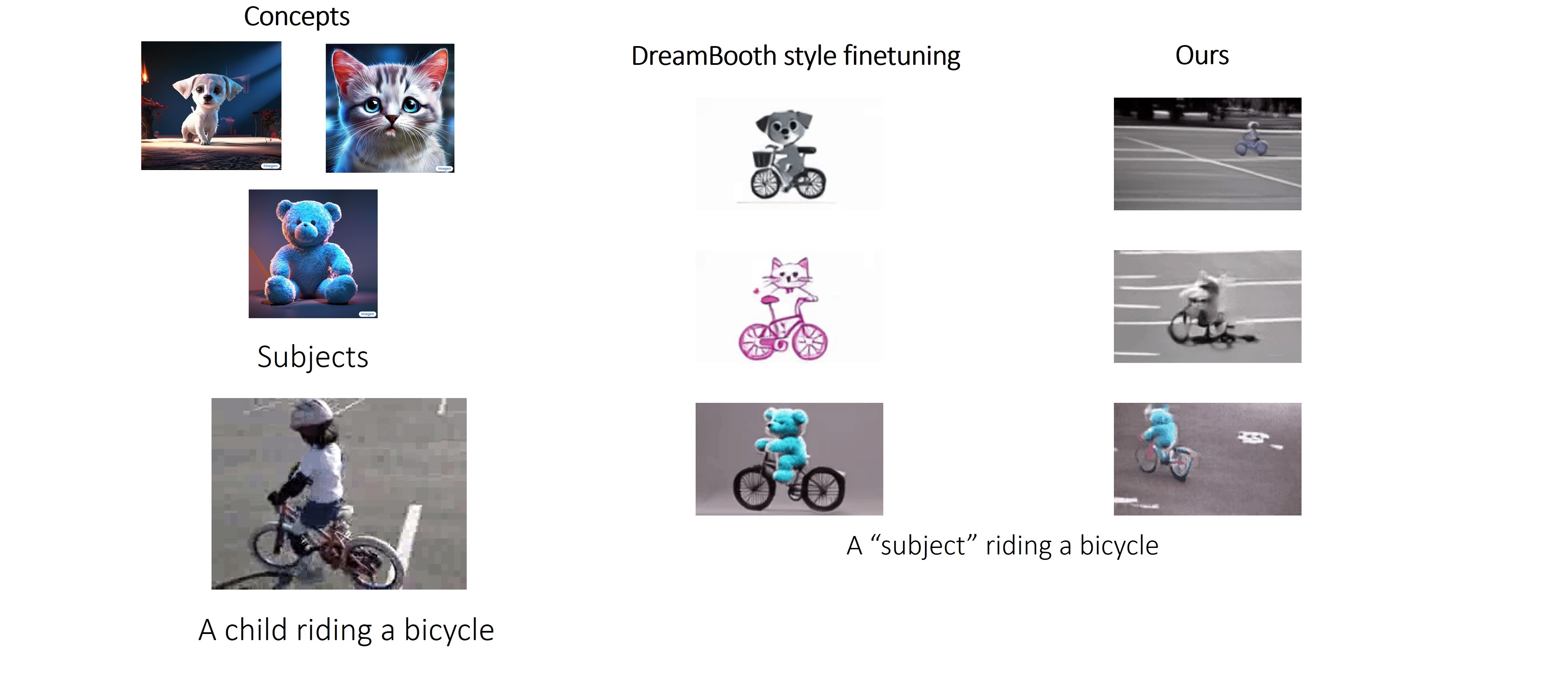}
    \caption{Additional results for subject-action customization.}
    \label{fig:supp7}
\end{figure*}

\begin{figure*}
    \centering
    \includegraphics[scale=0.5]{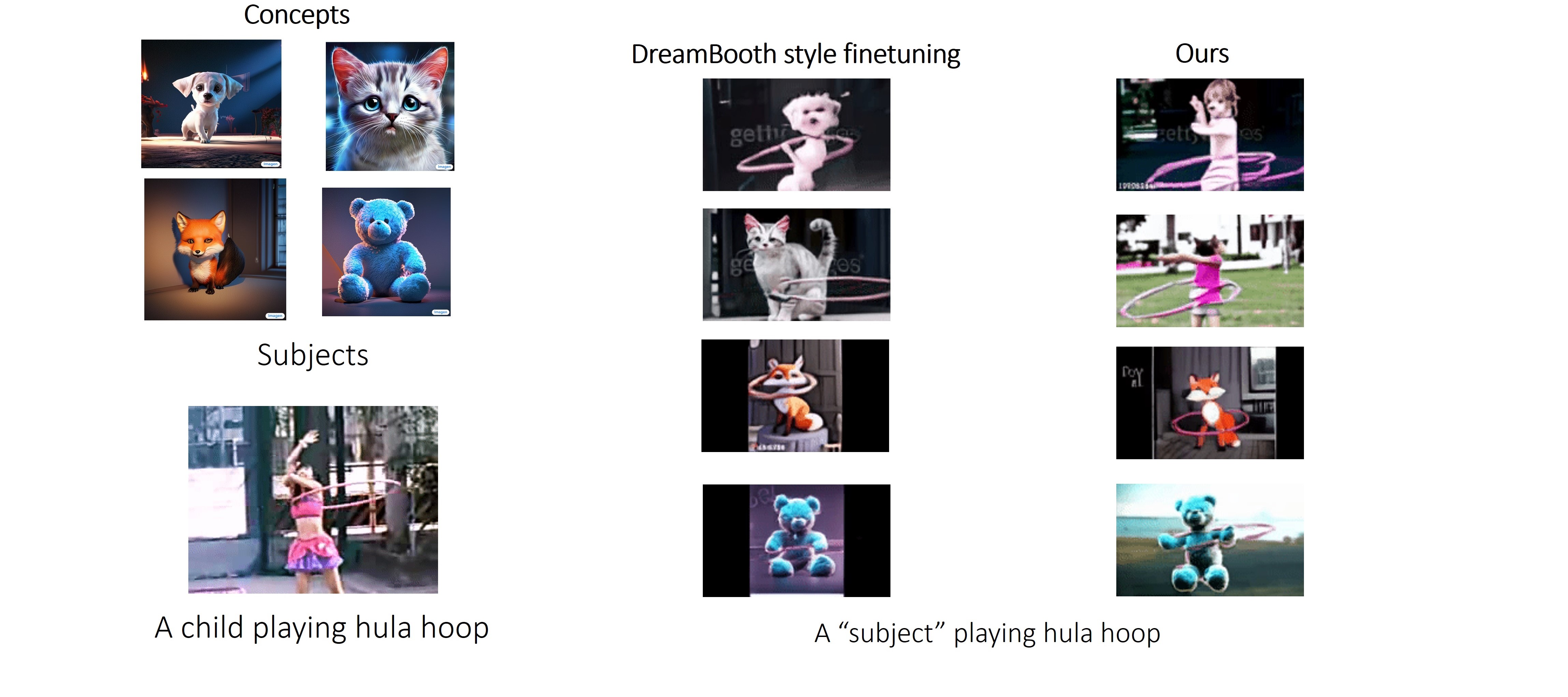}
    \caption{Additional results for subject-action customization. }
    \label{fig:supp8}
\end{figure*}

\begin{figure*}
    \centering
    \includegraphics[scale=0.5]{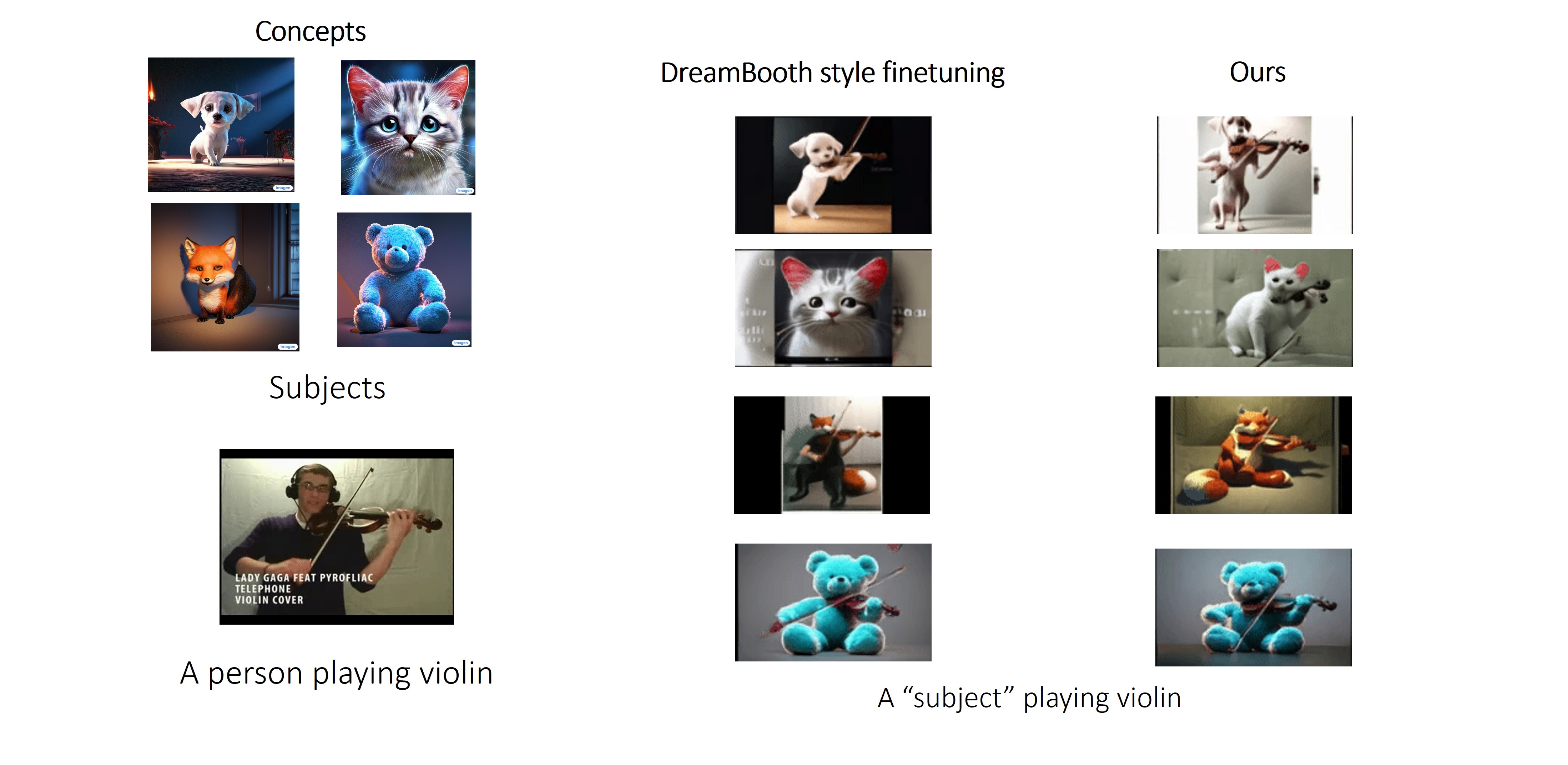}
    \caption{Additional results for subject-action customization. }
    \label{fig:supp9}
\end{figure*}

%%%%%%%%% REFERENCES
{\small
\bibliographystyle{ieee_fullname}
\bibliography{references}
}

\end{document}